# Language Hierarchization Provides the Optimal Solution to Human Working Memory Limits

Luyao Chen[1,2*], Weibo Gao[3], Junjie Wu[4,5,6,7], Jinshan Wu[8*], Angela D. Friederici[2*#]

[1]Department of Language Science and Technology, Faculty of Humanities, Hong Kong Polytechnic University, Hong Kong, China

[2]Department of Neuropsychology, Max Planck Institute for Human Cognitive and Brain Sciences, Leipzig, Germany

[3]State Key Laboratory of Cognitive Intelligence, University of Science and Technology of China, Hefei, China

[4]Key Research Base of Humanities and Social Sciences of the Ministry of Education, Academy of Psychology and Behavior, Tianjin Normal University, Tianjin, China;

[5]Faculty of Psychology, Tianjin Normal University, Tianjin, China.

[6]Department of Linguistics, Faculty of Medicine, Health and Human Sciences, Macquarie University, Sydney, NSW, Australia.

[7]Tianjin Key Laboratory of Student Mental Health and Intelligence Assessment, Tianjin, China

[8]Institute of Educational System Science, School of Systems Science, Beijing Normal University, Beijing, China

*Co-corresponding authors:

Prof. Dr. Luyao Chen

8 Hung Lok Road, Hung Hom, Kowloon, Hong Kong, China

harry-luyao.chen@polyu.edu.hk

Prof. Dr. Jinshan Wu

No. 19, Xinjiekouwai Str., Haidian Dist., Beijing, China

jinshanw@bnu.edu.cn

Prof. Dr. Angela D. Friederici

Stephan Straβe 1A, Leipzig, Germany

friederici@mpg.cbs.de

# Senior Author

**ABSTRACT**

Language is a uniquely human trait, conveying information efficiently by organizing word sequences in sentences into hierarchical structures[1-4]. A central question persists: Why is human language hierarchical? In this study, we show that hierarchization optimally solves the challenge of our limited working memory capacity[5-7]. We established a likelihood function that quantifies how well the average number of units according to the language processing mechanisms aligns with human working memory capacity (WMC) in a *direct* fashion. The maximum likelihood estimate (MLE) of this function, $\theta_{\text{MLE}}$, turns out to be the mean of units. Through computational simulations of symbol sequences and validation analyses of natural language sentences, we uncover that compared to linear processing, hierarchical processing far surpasses it in constraining the $\theta_{\text{MLE}}$ values under the human WMC limit, along with the increase of sequence/sentence length successfully. It also shows a converging pattern related to children's WMC development. These results suggest that constructing hierarchical structures optimizes the processing efficiency of sequential language input while staying within memory constraints, genuinely explaining the universal hierarchical nature of human language[8].

**Graphic abstract**

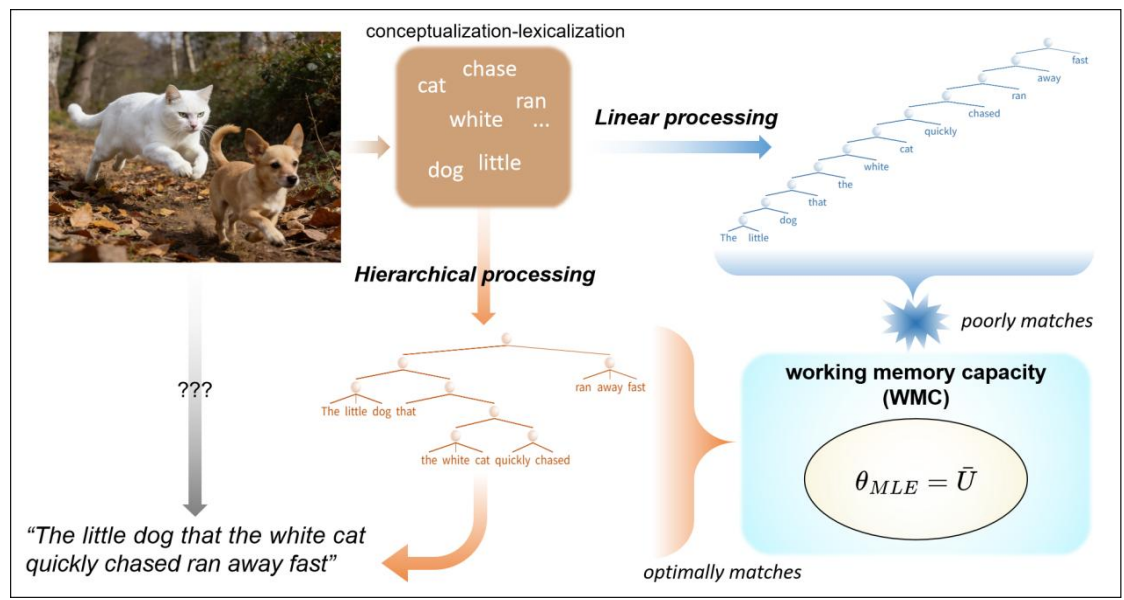

**MAIN**

Human communication primarily relies on language, enabling the efficient processing of large amounts of information in a short time. Upon receiving sequences of linguistic input, humans must rapidly decode language signals into speech sounds, words, phrases, and eventually full sentences[9]. As pinpointed by G. Miller[5] several decades ago, "Our language is tremendously useful for repackaging material into a few chunks rich in information," interpreting a sentence requires assembling its components into structural relationships, which in turn necessitates holding multiple elements in our limited working memory (WM)[10–15]. The developed (adult) human working memory capacity (WMC) is assumed to vary around 4 or 7 items[5-7]. For sentence lengths exceeding the human WMC limits, it would be unmanageable to process and interpret these sentences if each word were stored individually or in a linear fashion. Thus, to process longer sentences, it is necessary to reduce the effective number of words by integrating them into larger groups, thereby lowering working memory demands[16-17]. This process, known as "hierarchization," the hierarchical processing mechanism merging words into higher-order constituents for retaining processing efficiency[18]. For instance, the 12-word sentence "The little dog that the white cat quickly chased ran away fast." is grouped into constituents, such as a determiner phrase, "[the [white cat]]," a verb phrase, "[quickly chased]," and a relative clause "[that [[the [white cat]] [quickly chased]]]" (see also Fig. 1a). A recent study proposed that the hierarchical structures could be implemented using global and local ranks, which are separately encoded as dimensions spanning a two-dimensional space at the neural level for efficient maintenance of language information[19].

While the relationship between language hierarchization and general cognitive abilities like WMC has long been assumed, why and how such a mechanism emerges remains a puzzle[20]. Humans differ from non-human primates not only in the overall size of their WMC (2 ± 1 items in non-human primates[21]), but more importantly in their capacity for developing hierarchical language structures, a uniquely human language faculty[22-23]. Previous studies have revealed that humans pursue a "least effort" strategy when processing language, as reflected in certain language features, such as minimizing the dependency distance, which quantifies how far a word depends on another (as measured by the number of intervening words) [24-27]. A very recent study hypothesized that human language should be structured to minimize the complexity of sequential predictability, as measured by excess entropy[14]. However, a *direct* link between how human language is structured/processed and domain-general cognitive abilities has not yet been established. Specifically, a fundamental question remains unresolved: Does the hierarchization of human language emerge as the optimal solution for efficiently processing large volumes of information under humans' WMC constraints?

To this end, we first establish a likelihood function, that is, a probabilistic model which directly integrates linguistic features and working memory capacity (WMC), to test the hypothesis that language hierarchization represents the optimal solution to the human WMC constraints in a relatively broad yet empirically plausible range of sentence length[28]. The key body of this function is a Gaussian function, incorporating the exponent term that computes the squared standardized distance between the number of units acquired by counting the open nodes of each word according to the given structures (denoted as $u_i$ at the *i-th* word)[29] and an unknown WMC parameter, noted as $\theta$ to be estimated. "Open nodes" refer to the items to be merged during the dynamic process of sentence processing, actively maintained in WM[29]. The Gaussian function assumes that the difference between $u_i$ and $\theta$ follows a normal distribution. Through mathematical derivation, the maximum likelihood estimate (MLE) of this likelihood function, $\theta_{MLE}$, is the mean of units (See Fig. 1a and *Methods*). Therefore, larger $\theta_{MLE}$ directly corresponds to the higher WM load. As shown in Fig. 1a, processing "the little dog" requires $\theta_{MLE}$ of 2. It is noteworthy that since non-human primates' WMC is 2 ± 1 items[21], in the present study, human WMC was defined as ranging from 4 to 9 items, a deliberately broad range chosen to circumvent debates surrounding the exact "magic number" of WM[5-7]. Our primary objective is to investigate whether language hierarchization confers an advantage in mitigating the WM load (quantified by $\theta_{MLE}$) relative to the linear processing mechanism inherent to the branching structures. The simulations on abstract symbolic sequences reveal that language hierarchization is able to reduce the $\theta_{MLE}$ values to the human WMC range, when compared with the linear processing of branching structures. Further tests in natural language sentences, including the corpora of the classic works in English, the typologically different languages, and of the English child spoken sentences have validated these findings and demonstrate this fundamental principle underpinning language hierarchy.

**RESULTS**

Simulations were first performed to determine whether the hierarchical structures generated through merging the words into various multi-level constituents mimicking natural human language structures, that is, the "hierarchical processing mechanism," outperformed the linear processing mechanism of the branching structures, such as "[...[[[[$s_0$],$s_1$]$s_2$]$s_3$]...$s_n$]," which linearly combines one word to the previous ones in each time (Fig. 1b). The hierarchical structures were generated by defining two merge mechanisms (Fig. 1b): (a) The strict balanced binary merge, which demands a parent node to contain two child nodes at each level (unless there is only one child node left), such as "[[[$s_0$, $s_1$], [$s_2$, $s_3$]], [[$s_4$, $s_5$], [$s_6$, $s_7$]]]." (b) The loose multi-node merge, which requires a parent node to contain 1-4 child nodes randomly, such as, "[$s_0$, [[[$s_1$, $s_2$], [$s_3$, $s_4$, $s_5$]], [$s_6$, $s_7$, $s_8$]]]." The binary merge is considered the simplest mechanism to generate hierarchical structures[18, 22]. For each length of the sequences (ranging from 1 to 100 words in this study), 1,000 structure tokens were randomly generated by each mechanism. As shown in Fig. 1c, it is clear that: (a) Language hierarchization depending on both merge mechanisms outperforms the linear processing mechanism underlying the branching structures and minimizes the WMC loads even when the sequence length is extremely long; (b) Its $\theta_{MLE}$ grows in a logarithmic manner; (c) Within a plausible sentence length range (5-30 words)[28], the corresponding $\theta_{MLE}$ mainly occurs in the human WMC scope. Moreover, the Shannon entropy was calculated for each processing mechanism per sequence length, and both merge mechanisms showed lower entropy than the linear processing mechanism of the branching structures (Fig. 1d).

----- Inset Fig. 1 about here ----

Validation in natural languages further provides a consistent pattern of the $\theta_{MLE}$ distribution. Several validation steps were taken. In a first step, texts from nine randomly-selected classic works (written or translated in English) were taken for the initial validation. Second, in order to test the mechanism's generality, we performed a cross-language validation. We selected *Alice's Adventures in Wonderland,* written Lewis Carroll (1865), one of the most popular iconic children's classics in the world, and its translated versions in seven other languages, including Chinese, French, German, Russian, Japanese, Italian, and Spanish, in addition to English. Third, to explore the mechanism's validity for language development, a validation test in the children development scenario (raging from 3-10 years of age) was performed. The sentence structures of these natural language corpora were automatically analyzed for unit/open node counting by Stanza[30]. The validation test on the classics corpus revealed that language hierarchization minimized the $\theta_{MLE}$ within the human WMC when the sentence length increased (mean $\theta_{MLE}$ = 5), which is consistent with the findings of the simulations (Fig. 2a). Likewise, all the eight languages shared a similar $\theta_{MLE}$ distribution pattern, in which language hierarchization outperformed the linear processing mechanism (mean $\theta_{MLE}$ = 4) (Figs. 2b-c).

----- Inset Fig. 2 about here ----

Furthermroe, with respect to the "child spoken language corpus," validation results uncovered a consistent pattern for all the age groups (Figs. 3a-b), in which language hierarchization reduced the $\theta_{MLE}$, and the mean $\theta_{MLE}$ of each age group exhibited a development curve (nonlinear regression: $R^2$ = 0.746, $F$ = 14.657, $p$ < .05)(Fig. 3c), aligning well with the developing WMC trajectory[31].

----- Inset Fig. 3 about here ----

**DISCUSSION**

Just as profound physical laws underpin the mundane phenomenon of an apple falling to the ground, the ordinary daily human language communication is governed by inherent natural principles as well: Language enables us to fold linear sequential information into a hierarchical structure further used during comprehension and production. In this study, we propose the metric "$\theta_{MLE}$" to *directly* quantify the relationship between human language structures and WMC. We initially conduct both exploratory and confirmatory data simulations on artificial and natural language materials, respectively, and unravel a potential law of language with the novel empirical evidence: Language hierarchization constitutes a remarkable mechanism for human language, which is constrained by our limited WMC and optimally reduces the WM loads to a specific range when compared with the linear processing mechanism underlying the branching structures. We summarized two sub-laws of language hierarchization as follows:

(a) Efficiency: $\theta_{MLE}|_H \ll \theta_{MLE}|_L$ (i.e., hierarchical processing [H] leads to much less [$\ll$] WM loads than linear processing [L]);

(b) Ecological validity: $\theta_{MLE}|_H \subset_\approx$ human WMC (i.e., hierarchical processing primarily mitigates WM load, keeping it within the bounds of human WMC for sentences of reasonable lengths).

Previous studies have indicated associations between hierarchical processing and human WMC[17, 25], although the neural substrates underlying both abilities might be separable[34–38]. However, such results cannot account for a more fundamental question: Why do humans need hierarchical structures as well as the underlying hierarchical processing mechanism at all? Our tentative hypothesis is that to "pack up" more information into a processable sentence, we have to generate hierarchical structures to ease the WM burden. This aligns with the quotation from G. Miller[5] at the very beginning, but by taking a step further, we emphasize the importance of language hierarchization: Efficient language processing is particularly enabled by hierarchization under the constraints of limited human WMC. To test this hypothesis directly, we developed a likelihood function that quantifies the relationships between language structures and WMC. The simulation results consistently show that language hierarchization outperforms the linear processing mechanism to minimize the $\theta_{MLE}$ within the human WMC scope when the sentence length gets longer. Such results may further compensate the findings of the preference for shorter linear dependency distances[24-25, 27] and lower prediction effort/excess entropy[26], as these features are identified in the linear sequences which are actually externalized from the internal hierarchical structures[1]. In particular, language hierarchization enables the optimal mitigation of the sentence entropy so as to improve the processing efficiency (Fig. 1d). Thus, $\theta_{MLE}$ plays a critical role in relating language hierarchization with WMC directly to explain the tensions between the processing efficiency and the general cognitive (i.e., WMC here) constraints.

Validation in the natural language sentences across different languages revealed a consistent pattern. Language hierarhization in all the natural language corpora tested here is critical to minimize the values of $\theta_{MLE}$ and to keep them under the upper-limit of the human WMC scope[5-7]. Moreover, such a pattern is shared across the typologically-different languages, consistent with the findings of a recent study[39], which found that language hierarchical universals are among the most salient across human languages suggesting the mechanism's generality. The influence of linguistic typological parameters on inter-individual differences in $\theta_{MLE}$ values across these languages represents a compelling avenue for future research. Furthermore, as revealed by the validation results in the child spoken language corpus, the change in $\theta_{MLE}$ values is nicely corresponding to the child's WMC development trajectory[31], suggesting that the faculty of language hierarchization is present early and closely related to the development of children's WMC[40]. It is noteworthy that, the child $\theta_{MLE}$ significantly surpasses "3" after the age of 7 years (Fig. 3c), which is reminiscent of the syntactic ability changes proposed in previous studies[4]. Children are able to process syntactically complex sentences with deeper hierarchical embedding after the age of about 7[4]. More critically, the present analysis offers a rigorous simulation methodology to refute the hypothesis that children depend on linear processing mechanisms for spoken sentence generation. Unlike language hierarchization, treating the hierarchical structures as the mere branching structures for linear processing cannot delineate the actual WMC developmental trajectory[31].

In conclusion, this paper unravels that the hierarchical nature of human language is not arbitrary but an inevitable optimal solution in confront of the constraints on human WM. Through establishing the likelihood function and the derivation of the metric, "$\theta_{\text{MLE}}$," directly linking language processing mechanisms and working memory, our work consistently provides a universal mechanistic explanation for the inherent drive toward hierarchical organization in language, based on direct computational simulations in abstract sequences and natural sentence corpora. As a key outlook, any model of language evolution must now account for this "inevitability," suggesting that future research in neurocognitive science should consider this fundamental link as a central element in understanding the development and complexity of the unique human language faculty.

**METHODS**

**Establishing the likelihood function for the maximum likelihood estimate**

This function expresses the probabilistic relationships between the number of units to be processed and the working memory capacity (WMC, noted as "$\theta$"). If they are perfectly matched at a certain processing stage, say, 7 units vs WMC = 7, then the structure could be efficiently processed without the waste of the working memory resource (i.e., in the situation when the units to be processed are less than the WMC value) or the overburden of our WMC (i.e., in the situation that the units exceed the WMC limit). Thus, a perfect match at a certain stage is assumed to reflect the most economic/optimal processing efficiency. We first employed the open node counting approach[29] to traverse a hierarchical structure to count the number of units/open nodes at each word ($u_i$). Then, for each $u_i$, a Gaussian function was set up to ensure the result ranges from 0 to 1, with 1 indicating the perfect match with a given WMC:

$$e^{-\frac{1}{2}(\frac{u_i-\theta}{\sigma})^2}$$

Therefore, the likelihood function is defined as:

$$L(u|\theta) = \prod_{i}^{n} e^{-\frac{1}{2}(\frac{u_i-\theta}{\sigma})^2}$$

where "$n$" is the number of words. The maximum likelihood estimate (MLE) of $\theta$ is:

$$\theta_{MLE} = \bar{U} = \frac{1}{n} \sum_{i=1}^{n} u_i$$

Moreover, the Shannon entropy was computed (see also Fig. 1a). For each processing mechanism (balanced binary merge, left-branching, and random multi-node merge [1–4 nodes]), we first computed the open-node count for every word in a sequence using a hierarchical traversal algorithm that accounts for nested depth and sub-component offsets. From the resulting distribution of open-node values across words, we derived the probability mass function $p_i$ as the frequency of each unique open-node count normalized by sentence length. We then calculated the Shannon entropy (base 2, units: bits) as:

$$H = -\sum_{i} p_i \log_2(p_i)$$

where the sum excludes zero-probability terms to avoid numerical singularities. For non-deterministic structures (left-branching, multi-node merge), we averaged entropy values over 1,000 random instantiations to reduce sampling variance; the deterministic strict balanced binary merge required only a single computation per sentence length.

**Simulations in the artificial symbolic sequences**

Sequences composed of meaningless symbols (such as $s_1$, $s_2$, and so on to represent words) form a "sentence pool" for the subsequent simulation (the longest sequence contains 100 words). Note that these symbols are free of semantic information to avoid semantic facilitation, and only the randomly assigned structures for each sequence of a given length will be analyzed. As introduced in the main text, these structures include the branching structures and the hierarchical structures generated by the strict binary merge and the loose multi-node (1-4 nodes) merge mechanisms. For each sequence length, 1,000 structure tokens were generated. To note, the structure types of the loose multi-node merge will increase along with the sentence length, while the other two types will not change except their lengths. The structure tokens denote to the structures containing different tokens (symbols). For instance, "$[[[s_0]s_1]s_2]$" and "$[[[s_3]s_4]s_5]$" are two structure tokens, sharing the same structure type (i.e., the left branching structure of the three-word sentence). Therefore, both the strict binary merge and the left branching lead to the fixed structure types with little variance. Open node counts were computed by designating the outermost list as a common parent node, whose immediate elements were identified as direct sub-components; words in the first sub-component started at 1 and increment by 1 per nesting level, while those in subsequent sub-components added an offset equal to the number of preceding sub-components to their depth-based count (see also Fig. 1a for illustration). Thus, the global mean curves of the $\theta_{\text{MLE}}$ values across the sequence lengths were obtained.

**Validation tests in natural languages**

We validated the numerical simulation results in natural-language sentences. Firstly, 9 classic works (in English) were randomly selected from the Top 100 downloaded list of an online electronic book website (https://www.gutenberg.org/). These classic works include *Pride and Prejudice* (Jane Austen, 1813), *Little Women* (Louisa May Alcott, 1868), *Crime and Punishment* (Fyodor Dostoyevsky, 1866), *Jane Eyre* (Charlotte Brontë, 1847), *A Tale of Two Cities* (Charles Dickens, 1859), *The Great Gatsby* (F. Scott Fitzgerald, 1925), *The Adventures of Tom Sawyer* (Mark Twain, 1876), *War and Peace* (Leo Tolstoy, 1869), and *Great Expectations* (Charles Dickens, 1861). The global mean $\theta_{MLE}$ curves of both the linear processing mechanism which parses the sentences as the branching structures and of the hierarchical processing mechanism were then analyzed based on the "classics corpus." Secondly, in order to test the cross-language universality of our hypothesis, the famous children's classic, *Alice's Adventures in Wonderland* (Lewis Carroll, 1865), was selected[33]. In consideration of the copyrights of the translated versions and to unify the translation style, we asked the Deepseek API (version 3.2, non-thinking mode) to translate the original English work into the other 7 languages, including Chinese, French, German, Russian, Japanese, Italian, and Spanish, so as to compose an "Alice corpus." Based on this corpus, $\theta_{MLE}$ curves for each language and the global mean $\theta_{MLE}$ curve were computed with respect to the two processing mechanisms. Lastly, we also expect to detect changes in $\theta_{MLE}$ values in language development, in that children's WMC develops with age[32], and their language outputs should reflect their WMC (limits) at each year of age. The child English spoken sentences at each age group (such as 3 ≤ age < 4 years, noted as 3-4 Y) was randomly selected (3,000 per age group ranging from 3-10 years) from the CHILDES (https://talkbank.org/childes/) to form a "child spoken language corpus." The $\theta_{MLE}$ curve for each age level and the global mean $\theta_{MLE}$ curve were calculated, and the nonlinear regression was used to estimate the change of the $\theta_{MLE}$ values To note, natural sentences were cleaned according to the sentence length criterion for all the corpora: 25% score of the sentence length (*Q1*) - 1.5 IQR ≤ valid sentence length ≤ 75% score of the sentence length (*Q3*) + 1.5 IQR (IQR: "Inter-quartile Range" = *Q3* - *Q1*). Descriptive statistics for each corpus were presented in *SI* Table S1.

## ACKNOWLEDGEMENTS

We sincerely thank Prof. Liping Feng and Dr. Emiliano Zaccarella for their inspiring discussions and constructive inputs into this work.

## AUTHOR CONTRIBUTIONS

**Luyao Chen:** Conceptualization, Formal Analysis, Software, Writing – Original Draft Preparation, Writing – Review & Editing, & Funding Acquisition; **Weibo Gao:** Formal Analysis, Software, & Writing – Original Draft Preparation. **Junjie Wu:** Writing – Review & Editing. **Jinshan Wu:** Conceptualization, Formal Analysis, & Writing – Review & Editing. **Angela D. Friederici:** Conceptualization, Formal Analysis, Software, Writing – Original Draft Preparation, and Writing – Review & Editing.

## COMPETING INTERESTS

The authors declared no conflict of interest in this work.

## FUNDING

This work was supported by STI 2030—Major Projects+2021ZD0200500.

## DATA AND CODE AVAILABILITY

All computational models, and the code for the analyses are available and will be published on our GitHub account. The mini-corpora will be available upon reasonable requests.

**Figure legends**

**Fig. 1. Numerical simulation results.**

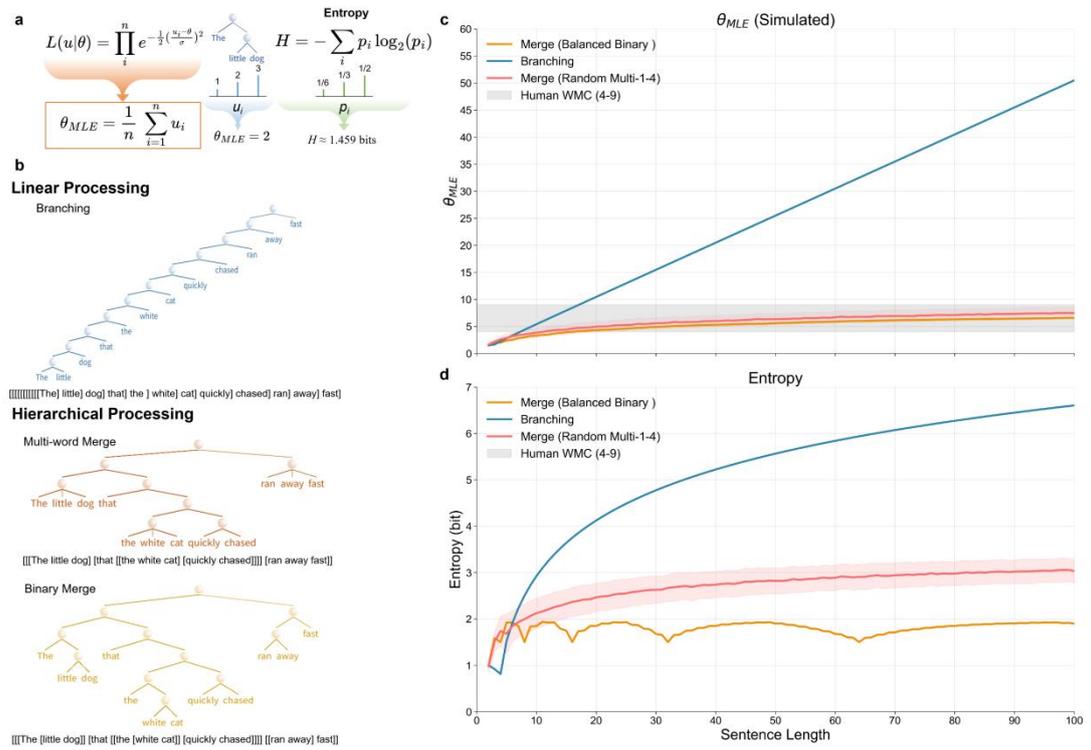

**a.** The core formulae adopted in this study. An example of calculating $\theta_{\text{MLE}}$ and the entropy ($H$) was provided. **b.** Illustration of the linear and hierarchical processing mechanisms. **c.** The simulation results of $\theta_{\text{MLE}}$. **d.** The simulation results of entropy.

**Fig. 2. Validation results in the written natural language corpora.**

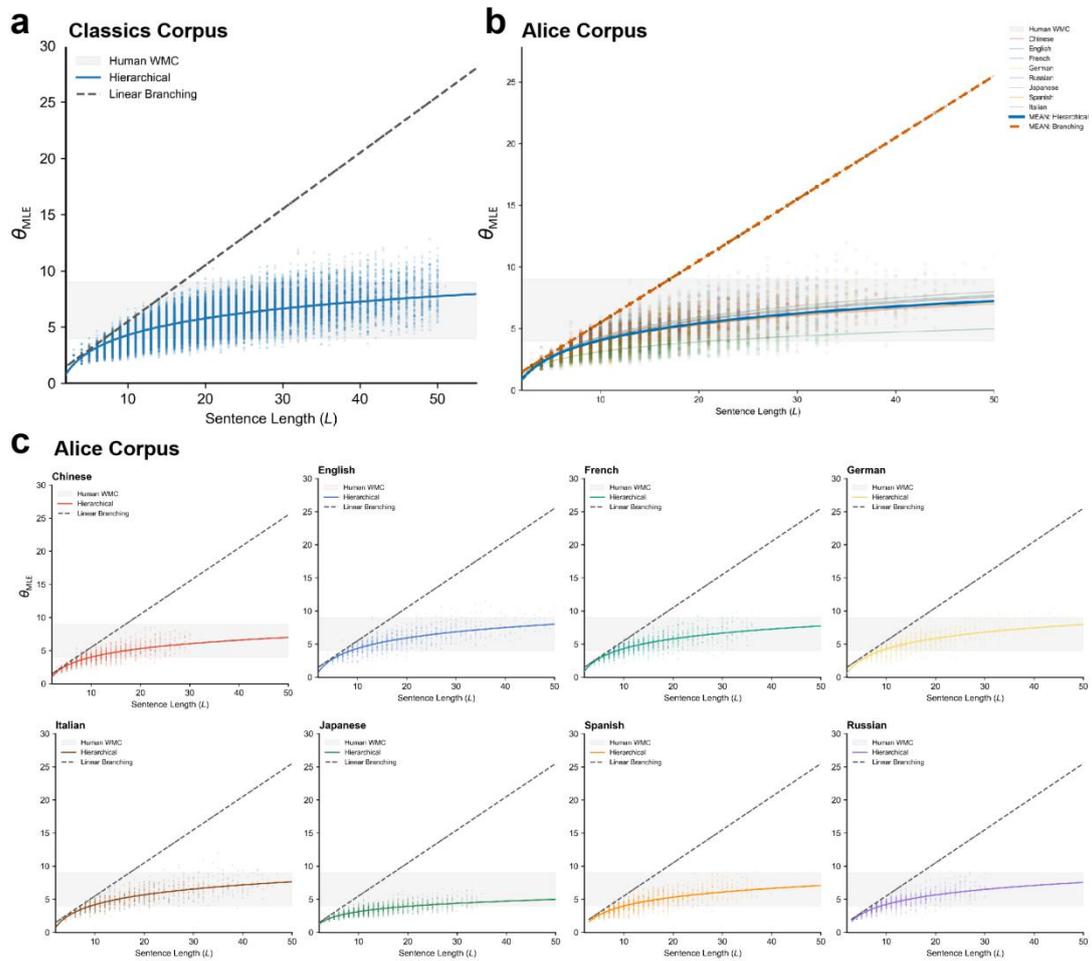

**a.** Validation test results in the "classics corpus." **b.** Validation test results (the global mean of all languages) in the "Alice corpus." **c.** Simulation results in the "Alice corpus." for each language.

**Fig 3. Validation results in the children spoken language corpus**

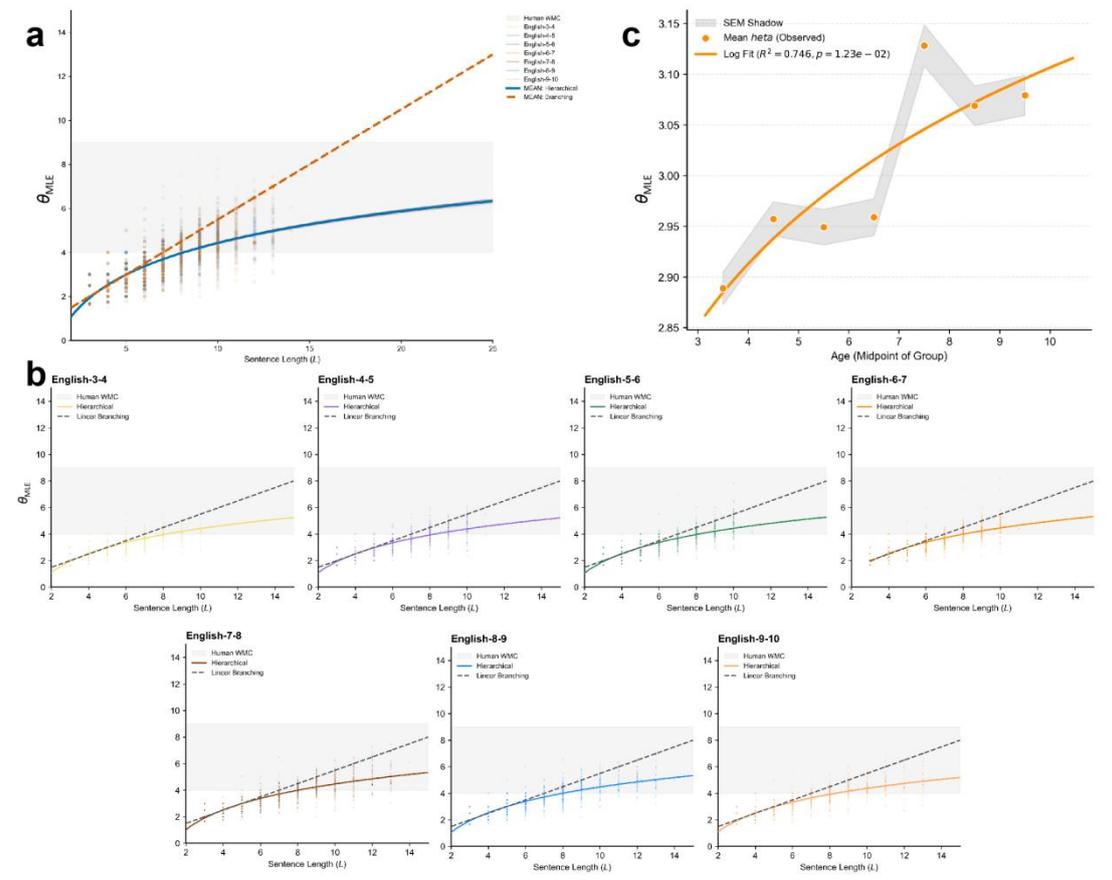

**a.** Validation test results (the global mean of all age groups) in the "child spoken language corpus." **b.** Validation test results in the "child spoken language corpus." for each age group. **c.** Nonlinear regression results between $θ_{MLE}$ and the age groups.